%% file: main.tex
\documentclass[conference]{IEEEtran}
\IEEEoverridecommandlockouts

\input{header}

\begin{document}

    \title{
        Towards Probabilistic Clearance, Explanation and Optimization
    }
    
    \author{
        Simon Kohaut$^{1, *}$, Benedict Flade$^{2, *}$, Devendra Singh Dhami$^{3, 4}$, Julian Eggert$^{2}$,  Kristian Kersting$^{1, 4, 5, 6}$
        \thanks{
            $^{*}$ Authors contributed equally
        }
        \thanks{
            $^{1}$ Department of Computer Science,\newline\hspace*{1.6em}
            TU Darmstadt, 64283 Darmstadt, Germany \newline\hspace*{1.6em}
            {\tt\small firstname.surname@cs.tu-darmstadt.de}%
        }%
        \thanks{
            $^{2}$ Honda Research Institute Europe GmbH, \newline\hspace*{1.6em} 
            Carl-Legien-Str. 30, 63073 Offenbach, Germany \newline\hspace*{1.6em}
            {\tt\small firstname.surname@honda-ri.de}
        }%
        \thanks{
            $^{3}$
            Uncertainty in Artificial Intelligence Group, \newline\hspace*{1.6em}
            Department of Mathematics and Computer Science, \newline\hspace*{1.6em}
            TU Eindhoven, 5600 MB Eindhoven, Netherlands%
        }%
        \thanks{
            $^{4}$ Hessian AI
        }%
        \thanks{
            $^{5}$ Centre for Cognitive Science
        }%
        \thanks{
            $^{6}$ German Center for Artificial Intelligence (DFKI)
        }%
    }
    
    \maketitle
    
    \input{content/abstract}
    \input{content/introduction}
    \input{content/related_work}
    \input{content/methods}
    \input{content/results}
    \input{content/conclusion}

    \bibliographystyle{IEEEtran}
    \bibliography{references.bib}

\end{document}

%% file: header.tex
\usepackage{cite}
\usepackage{amsmath,amssymb,amsfonts}
\usepackage{algorithmic}
\usepackage{graphicx}
\usepackage{textcomp}
\usepackage[dvipsnames]{xcolor}
\usepackage{balance}
\usepackage{epstopdf}
\usepackage[normalem]{ulem}
\usepackage{upgreek}
\usepackage{comment}
\usepackage{url}
\usepackage{tikz}
\usepackage{lipsum}
\usepackage[formats]{listings}
\usepackage{subcaption}
\usepackage{multirow}
\usepackage{siunitx}
\usepackage[frozencache,cachedir=.]{minted}
\usepackage{tikz}
\usepackage[color=black,opacity=1,angle=0,scale=1]{background}
\backgroundsetup{
  contents={
    \begin{tikzpicture}
        \node at (current page.center) [align=center] {\textcopyright 2024 IEEE. Personal use of this material is permitted. \\ Permission from IEEE must be obtained for all other uses, in any current or future media, including reprinting/republishing this material for advertising or promotional purposes, \\ creating new collective works, for resale or redistribution to servers or lists, or reuse of any copyrighted component of this work in other works.
         \\ DOI: 10.1109/ICUAS60882.2024.10556879};
    \end{tikzpicture}},
  placement=bottom,
  scale=0.6,
  vshift=20
}

%% file: content/abstract.tex
Employing Unmanned Aircraft Systems (UAS) beyond visual line of sight (BVLOS) is an endearing and challenging task.
While UAS have the potential to significantly enhance today's logistics and emergency response capabilities, unmanned flying objects above the heads of unprotected pedestrians induce similarly significant safety risks.
In this work, we make strides towards improved safety and legal compliance in applying UAS in two ways.
First, we demonstrate navigation within the Probabilistic Mission Design (ProMis) framework.
To this end, our approach translates Probabilistic Mission Landscapes (PML) into a navigation graph and derives a cost from the probability of complying with all underlying constraints.
Second, we introduce the clearance, explanation, and optimization (CEO) cycle on top of ProMis by leveraging the declaratively encoded domain knowledge, legal requirements, and safety assertions to guide the mission design process.
Based on inaccurate, crowd-sourced map data and a synthetic scenario, we illustrate the application and utility of our methods in UAS navigation.


%% file: content/introduction.tex
\section{Introduction}
\label{sec:introduction}

\input{figures/motivation}

\PARstart{M}{ission} design is integral to the successful and safe execution of UAS missions. 
It involves achieving mission objectives while adhering to regulatory constraints and personal preferences.
Examples of such missions include parcel delivery~\cite{Huang2021}, search and rescue operations~\cite{Jayalath2021}, smart farming~\cite{Maddikunta2021}, or aerial photography~\cite{Wu2020}.

As such applications depend on navigation systems that reliably stick to predefined rules, we have previously presented the Probabilistic Mission Design (ProMis) framework \cite{Kohaut2023}.
In this work, we extend our prior work by tackling the following tasks in a probabilistic setting: 

\begin{itemize}
   \item Mission Clearance (C): Discriminating between missions that fulfill predefined rules and those that do not.
   \item Mission Explanation (E): Demonstrating how the parameters influence the decision of mission clearance.
   \item Mission Optimization (O): Searching and adapting parameters to improve mission performance concerning its underlying constraints.
\end{itemize}

As illustrated in Figure~\ref{fig:motivation}, the three tasks form a cyclical process, each informing the others. 
If a planned mission fails clearance, its critical parameters are identified in the explanation step, guiding optimization efforts. 
However, in order to streamline the mission design process, multiple challenges need to be addressed.
The first is safety.
Ensuring the reliability of UAS operations is a multifaceted task involving the safety of ground and airborne entities. 
Second, regulatory compliance poses a significant challenge. 
Using drones for private and commercial purposes is subject to numerous regulatory barriers, especially in novel or complex environments with varying yet-to-be-decided regulations. 
Third, public acceptance is crucial, including concerns such as privacy infringement, noise pollution, or the potential for accidents. 
The public hesitates to accept using uninterpretable black-box models to make critical decisions, as they challenge understanding and evaluating the decision-making process.
Addressing these challenges is essential for successfully implementing UAS missions and the broader integration of drones into society.
It is crucial to address these issues by moving away from black-box approaches and towards transparent systems with intuitive interfaces. 
Symbolic, white-box models offer a solution, such as probabilistic logic programs~\cite{problog,inference_in_plp}. 
These models represent rules and uncertainties in an interpretable and adaptable manner.

To this end, we have presented ProMis, a transparent system utilizing probabilistic logic to improve UAS navigation~\cite{Kohaut2023}. 
Probabilistic logic combines probability theory with logical reasoning, such as first-order logic, to model uncertainty and allow inferences based on probabilistic principles. 
ProMis provides a framework for representing uncertain navigational knowledge and constraint in a hybrid probabilistic setting, i.e., combining continuous and discrete distributions into a single model.
Hence, it leverages probabilistic reasoning to model uncertainty and to provide human-interpretable outputs while considering legal constraints. 
The elegance of this approach lies in the ability to model uncertainties or vague statements symbolically, making them easily understandable and adaptable afterward.
For example, translating inaccurate crowd-sourced maps into probabilistic logic captures their inherent uncertainty and improves the robustness and safety of mission design.

In this paper, we follow up on ProMis by making the following contributions to the UAS community:
\begin{itemize}
    \item We demonstrate path planning over ProMis' generated Probabilistic Mission Landscapes (PML) by translating them into a navigation graph with weighted edges. 
    \item The CEO cycle over ProMis, facilitating clearance, explanation, and optimization of trajectories grounded in PMLs for teleoperated or autonomous vehicles.
    \item An Open Source implementation of our methods as grounds for future application development, found at \url{https://github.com/HRI-EU/ProMis}.
\end{itemize} 

We first present related work connected to our contributions and recent developments in Advanced Aerial Mobility (AAM) and Unmanned Traffic Management (UTM).
Second, we present ProMis as the foundation of the CEO cycle by summarizing our prior work.
Third, we show the operations of a UAS on top of ProMis in the form of clearance, explanation, and optimization as individual building blocks for advancing safe and reliable aerial mobility.
Fourth, our results demonstrate the contributed building blocks in unison to demonstrate their applicability and utility.
Finally, we summarize this work and point towards future extensions.

%% file: figures/motivation.tex
\begin{figure}
    \centering
    \includegraphics[width=0.65\linewidth]{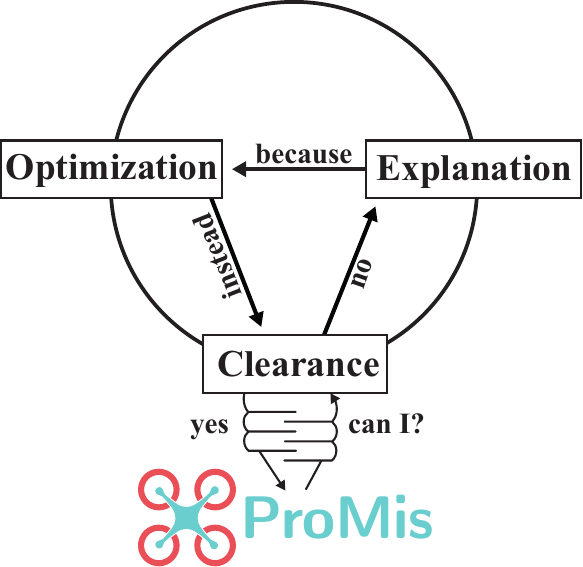}
    \caption{
        \textbf{The probabilistic CEO cycle:}
        If an intended Probabilistic Mission (ProMis) does not pass the clearance (C) step, explanation (E) identifies critical parameters that are passed to an optimizer (O) choosing the ideal setting, e.g., consisting of departure time, operator attributes, or UAS type.
        This cycle is transparent to the user, only acting on inputs likely to violate legal and safety constraints or operator preferences.
    }
    \label{fig:motivation}
\end{figure}

%% file: content/related_work.tex
\section{Related Work}
\label{sec:related_work}

In Europe, the Single European Sky Air Traffic Management Research (SESAR) initiative has significantly influenced the advancement of UAS and Urban Air Mobility (UAM). 
When SESAR introduced its first Master Plan in 2009, the concept of AAM was not considered, with drones merely mentioned in passing.
However, with subsequent editions of the Master Plan, particularly the 2015 release~\cite{undertaking2016european} and the 2017 Drone Outlook Study~\cite{sesar_2017}, the integration of UAS into European airspace gained momentum.

Besides substantial funding allocated to drive development, regulations, and operational restrictions have been introduced concurrently to ensure the safety of drone operations.
Notably, the EU regulation 2019/947~\cite{Ec2019}, implemented in 2019, underscores the coexistence of unmanned and manned aircraft within shared airspace, emphasizing rigorous risk assessments, such as through the Specified Operations Risk Assessment (SORA method)~\cite{Easa2019}.

Recent research endeavors have enhanced safety analysis and risk assessment methodologies. 
For instance, Rothwell and Patzek~\cite{RothwellPatzek2019} have contributed by employing satisfiability checks on symbolic models to verify and improve mission planning for UASs. 
In contrast, Rakotonarivo et al. \cite{Rakotonarivo2022} highlight the importance of directly fitting the output of safety analysis and risk assessment to map or environmental data. 
More specifically, they propose to improve the visual representation and enable data exploration by displaying interactive representations of mission parameters. 
This approach improves integration with environmental data, positively impacting the effectiveness of safety analysis and risk assessment in real-world scenarios.

Several studies, such as~\cite{Primatesta2020} and~\cite{Raballand2021}, have introduced approaches that integrate visual maps with risk models to calculate and visually represent risks associated with UAS operations. 
These models prioritize risk assessment based on formal frameworks, aiding in identifying potential hazards and safety concerns, especially ground casualties and transportation network disruptions.

We have previously presented the Probabilistic Mission (ProMis) framework \cite{Kohaut2023}.
It aligns with preceding research efforts, offering a formal symbolic approach to probabilistic classification and mission design. 
Notably, ProMis provides flexibility in mission design while allowing arbitrary queries related to mission parameters, thus facilitating comprehensive risk assessment and decision-making.
Moreover, ProMis accommodates evolving regulatory frameworks and operational priorities within its probabilistic framework, such as battery lifetime~\cite{An2023} or weather-effects~\cite{Schuchardt2022}.

Before concluding this section, we must note that we have yet to show how ProMis can generate trajectories, which will be a central discussion of this work.
Based on the ProMis-derived Probabilistic Mission Landscape (PML), we will illustrate how a path planner can generate or optimize paths using these PMLs.
Hence, path planning techniques, e.g., as introduced by \cite{Hohmann2022, Shen2022, Cabral2023, Papaioannou2021}, can complement ProMis by consuming the PML's data to guide trajectory generation towards behavior that complies with the underlying mission constraints. 
An example illustrating how a path planner utilizes a landscape (in this case, risk-based) is shown in \cite{Hu2020}. 
Similarly, we will show a path-planning approach in ProMis, albeit with the distinction that the underlying landscape can represent arbitrary mission-related queries besides purely collision-based factors.

This paper aims to elucidate the developed concepts and demonstrate their integration in a Clearance, Explanation, Optimization (CEO) cycle. 
By showcasing the potential of probabilistic logic, mainly through the ProMis framework, this work underscores its contribution to safety in UAS operations.
Besides, it complements our original publication by demonstrating ProMis' application by leveraging PMLs to guide navigation and each CEO step.

%% file: content/methods.tex
\section{Probabilistic Mission Design}
\label{sec:promis}

\input{figures/landscapes}
Let us briefly examine the basics of Probabilistic Missions (ProMis).
In ProMis, we combine mission constraints, domain knowledge, and operation details using Hybrid Probabilistic Logic Programs (HPLP).
HPLPs intertwine formal logic with probability theory over hybrid relational spaces, i.e., categorical and continuous distributions~\cite{nittihybrid, kumar2022first}.
To do so, an HPLP contains so-called distributional clauses (DC).
Hence, logical propositions can be either Boolean (true or false) or follow discrete or continuous distributions to parameterize the probability of a solution to the modeled problem.
In other words, one of the following can be used to express the fact that $\mathcal{A}$ is true with probability $p$ or its value follows the parameterized distribution $p(\mathbf{\theta})$.
\begin{align*}
    p &:: \mathcal{A} \leftarrow l_1, \ldots, l_n. \tag*{(Discrete)} \\
    \mathcal{A} &\sim p(\mathbf{\theta}) \leftarrow l_1, \ldots, l_n. \tag*{(Continuous)}
    \label{eq:distributional_clauses}
\end{align*}%
With this, the right-hand-side conjunction of literals $l_i$ is a precondition for atom $\mathcal{A}$ to follow the given distribution.

\input{listings/promis}

This formalization of knowledge in uncertain domains is especially interesting when considering crowd-sourced maps and low-cost sensory equipment.
Since creating and maintaining high-definition maps is an intractable, expensive task, creating methods incorporating less accurate and reliable data is invaluable to conquering advanced autonomy in any form of mobility.
For this reason, let us consider two types of geographic errors by applying a stochastic model.
Consider a map $M = (\mathcal{V}_M, \mathcal{E}_M)$ as collection of geo-referenced structures represented by vertices $\mathcal{V}_M$ and edges $\mathcal{E}_M$.
To obtain an HPLP representation of the underlying data, ProMis requires a generative model of map features to extract parameters of probabilistic spatial relations.
We utilize an affine transformation to model the spatial error in the stored vertices, i.e., a translation and a linear map.
Analogous to prior work \cite{flade2021error}, we consider for each $\mathbf{v}_{i, j} \in \mathcal{V}_M$, being the $j$-th vertex of the $i$-th feature, the following affine map:
\begin{align*}
    \mathbf{\Phi}^n &\leftarrow \phi(\mathbf{\alpha}_i) \tag*{(Transformation)} \\
    \mathbf{t}^n &\leftarrow \kappa(\mathbf{\beta}_i) \tag*{(Translation)} \\
    \mathbf{v}^n_{i,j} &= \mathbf{\Phi}^n \cdot \mathbf{v}_{i, j} + \mathbf{t}^n \tag*{(Generation)}
\end{align*}
Here, $\phi(\mathbf{\alpha}_i)$ and $\kappa(\mathbf{\beta}_i)$ are parameterized generators of linear maps $\mathbf{\Phi}$ and translations $\mathbf{t}$.
Obtaining a collection of $N$ samples then allows for computing, e.g., \textit{distance} and \textit{over} relations as a continuous distribution over $\mathbb{R}^+$ and probability $p \in [0, 1]$ respectively.
Figure~\ref{fig:pml_matrix} visualizes the resulting parameters over a navigation space.
Further, incorporating domain knowledge into the HPLP, e.g., laws of the local authorities, yields a model as presented in Listing~\ref{listing:uam_model}.
Querying this model across the navigation space produces a so-called Probabilistic Mission Landscape (PML), a scalar field indicating the probability of all constraints being met at the respective point.
A PML, as shown in Figure~\ref{fig:pml_matrix}, will be the basis for the following discussions.

\input{figures/architecture}

\section{The Probabilistic UAS}

We are now introducing fundamental concepts of probabilistic UAS operations leveraging the ProMis framework.
To do so, \textbf{C}learance, \textbf{E}xplanation, and \textbf{O}ptimization (CEO) over ProMis data are presented individually in the following sections.
Figure~\ref{fig:architecture} illustrates the overall idea of our approach:
While ProMis is a basis for generating Probabilistic Mission Landscapes (PML) as described in Section~\ref{sec:promis}, we show how the information of the PML and HPLP allows setting up, discriminating, and refining missions.

\subsection{Clearance}

From either operator, agent, or an optimization step as outlined later, we obtain a mission $\tau$.
In our setting, the mission $\tau$ carries not only the information of the via-points of the agent, i.e., its location at various points in time but further semantic labels that describe the (predicted or intended) state of the mission.
For example, $\tau_n = (x, y, \gamma, l, t)$ may describe Cartesian coordinates $x, y$ and yaw $\gamma$ at step $n$ in addition to the pilot's license $l$ and the time of day $t$.
Hence, each $\tau_n$ contains data for which the employed HPLP can compute the probability of complying with all navigation constraints, which, in turn, is stored in the PML.

The task of clearance is to assign a label to $\tau$ to discriminate whether $\tau$ fulfills all mission requirements, e.g., legality, safety, or operator constraints.
Let $P_L(\tau_n)$ be the probability of a semantic via-point being legal according to a PML $L$, as described in Section~\ref{sec:promis}.
Hence, one can compute the cost $J$ of a trajectory $\tau$ as the average probability of violating the PML along the way, i.e.,
\begin{equation}
    \label{eq:cost}
    J = \frac{1}{N} \sum_n 1 - P_L(\tau_n). 
\end{equation}
Given a threshold $T_J$, we decide for a trajectory $\tau$ to get clearance \textit{iff} $J < T_J$, i.e., a trajectory $\tau$ is valid if the average probability of violating $L$ is less than the set threshold.

To find the route of minimum cost over a given PML, one can transform the PML into a directed graph $G = (\mathcal{V}_G, \mathcal{E}_G)$.
A PML as a grid of samples of the HPLP becomes the set of vertices $\mathcal{V}_G$ while the probability of the respective node is taken as the weight of the edge pointing towards it.
Due to the normalization, long journeys can offset traversing high-cost edges.
Thus, a threshold $T_P$ can be applied to filter for such edges and avoid violations of the navigation constraints.

Applying a graph-based, shortest path search using these weights, e.g., Dijkstra, leads to a path through the PML with the minimum overall cost, which can then be normalized for clearance as shown in Equation~\ref{eq:cost}.

\subsection{Explanation}

Explaining (probabilistic) models aims to understand how inputs correspond to outputs.
A model explainer approximates the model's behavior, highlighting how each input pushes the output towards a label.
In the case of Clearance in ProMis, we can gain insights into which parameters of the scenario lead to a route being rejected or accepted.
In other words, one does not want to get a definitive answer to a query but to understand why ProMis generated the answer and how to best sway the setting for clearance.

Concerning the previously presented clearance, one might ask the question \textit{why} a desired trajectory was denied in ProMis.
In this case, we analyze the influence of the semantic parameters attached to the UAS path through the PML.
These parameters are not generated from geospatial data, indexed in Cartesian coordinates, but rather describe the mission and its circumstances on a higher level.
For example, a basic license for piloting UAVs can cause the denial of a mission's clearance in a demanding setting.
An explanation then aims to unveil the strong effect of the choice in license to inform further decision-making.
This explanation is obtained by creating variations of these parameters and reporting their effect on the outcome, i.e., how optimal routes differ when searching in the respective PML.

\input{figures/ceo}
\input{listings/overpass}
\subsection{Optimization}

While clearance informs about the acceptance of a proposed mission plan and explanation gives local insight into the impact of the mission parameters, searching for the optimal strategy allows the exploitation of free parameter choices to find the most compliant setting.
Given a start and goal position, we can search for a setting that minimizes the cost of reaching the goal.
In other words, we search for free parameters of the PML to find the trajectory and parameters pair with minimum cost $J$.
While geospatial data induces a majority of parameters, e.g., distributional clauses such as \textit{distance} and \textit{over}, a mission is defined through choices such as the time at which operations take place or the legal setting, such as employed license or type of UAS.
Hence, these choices on the mission framework are grounds for optimization, i.e., searching through the space of valid choices to procure a setting with minimal cost.
Section~\ref{sec:results} will illustrate how the complete CEO cycle can be applied to showcase its utility in a practical example.

%% file: figures/landscapes.tex
\begin{figure}
    \centering
    \begin{subfigure}{0.39\linewidth}
        \includegraphics[width=\textwidth]{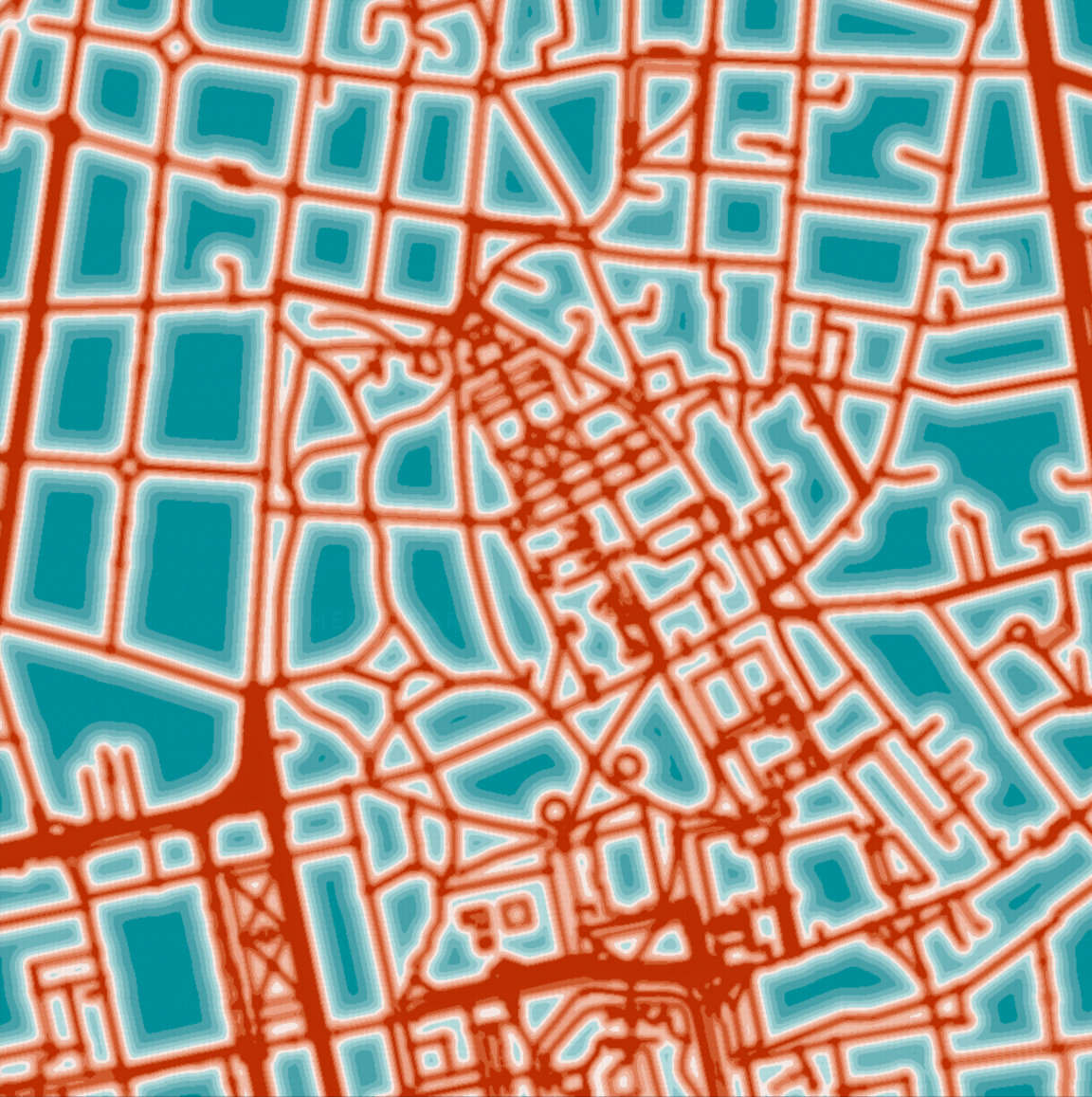}
    \end{subfigure} 
    \begin{subfigure}{0.39\linewidth}
        \includegraphics[width=\textwidth]{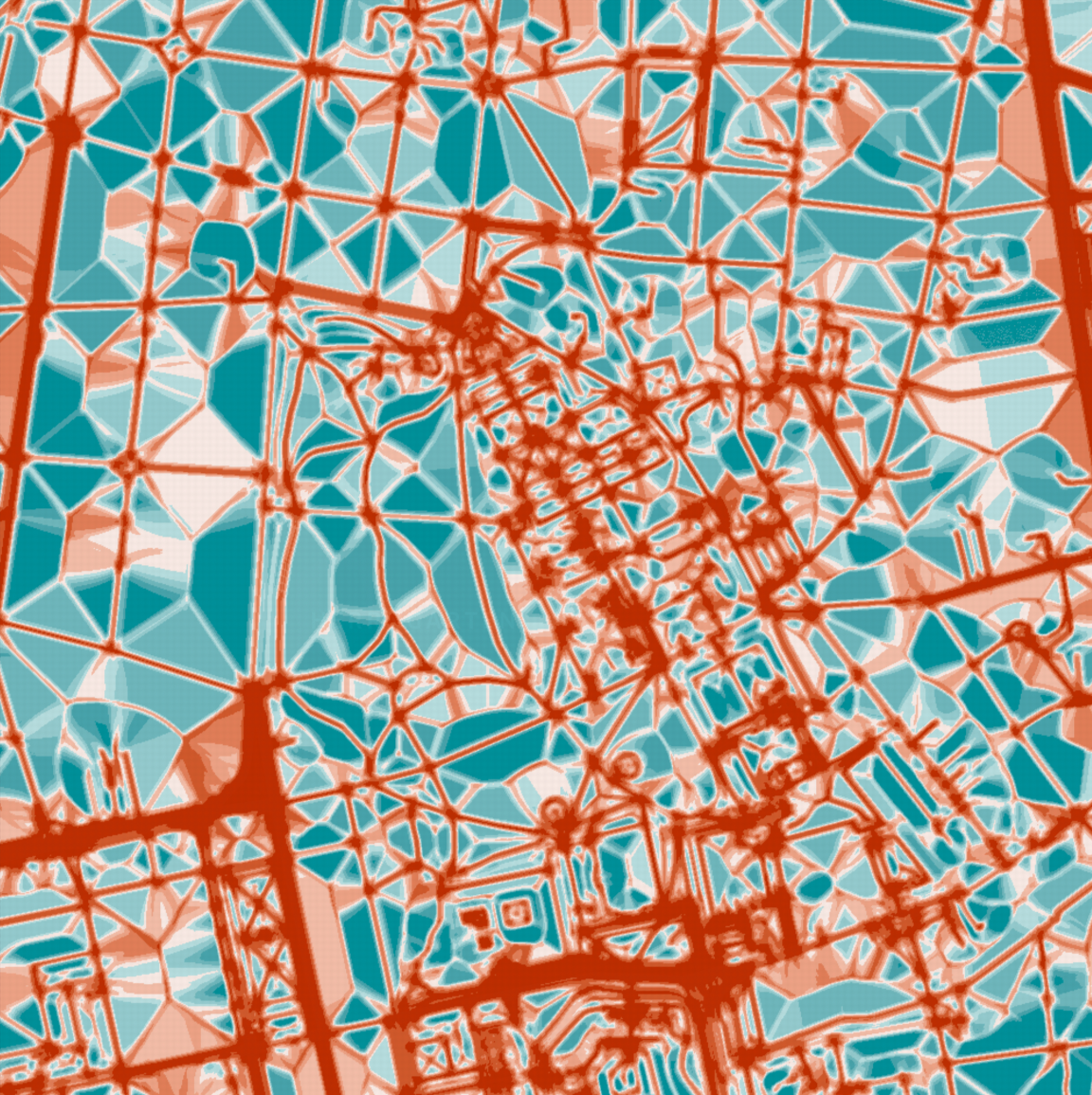}
    \end{subfigure} \\
    \vspace{0.35em}
    \begin{subfigure}{0.39\linewidth}
        \includegraphics[width=\textwidth]{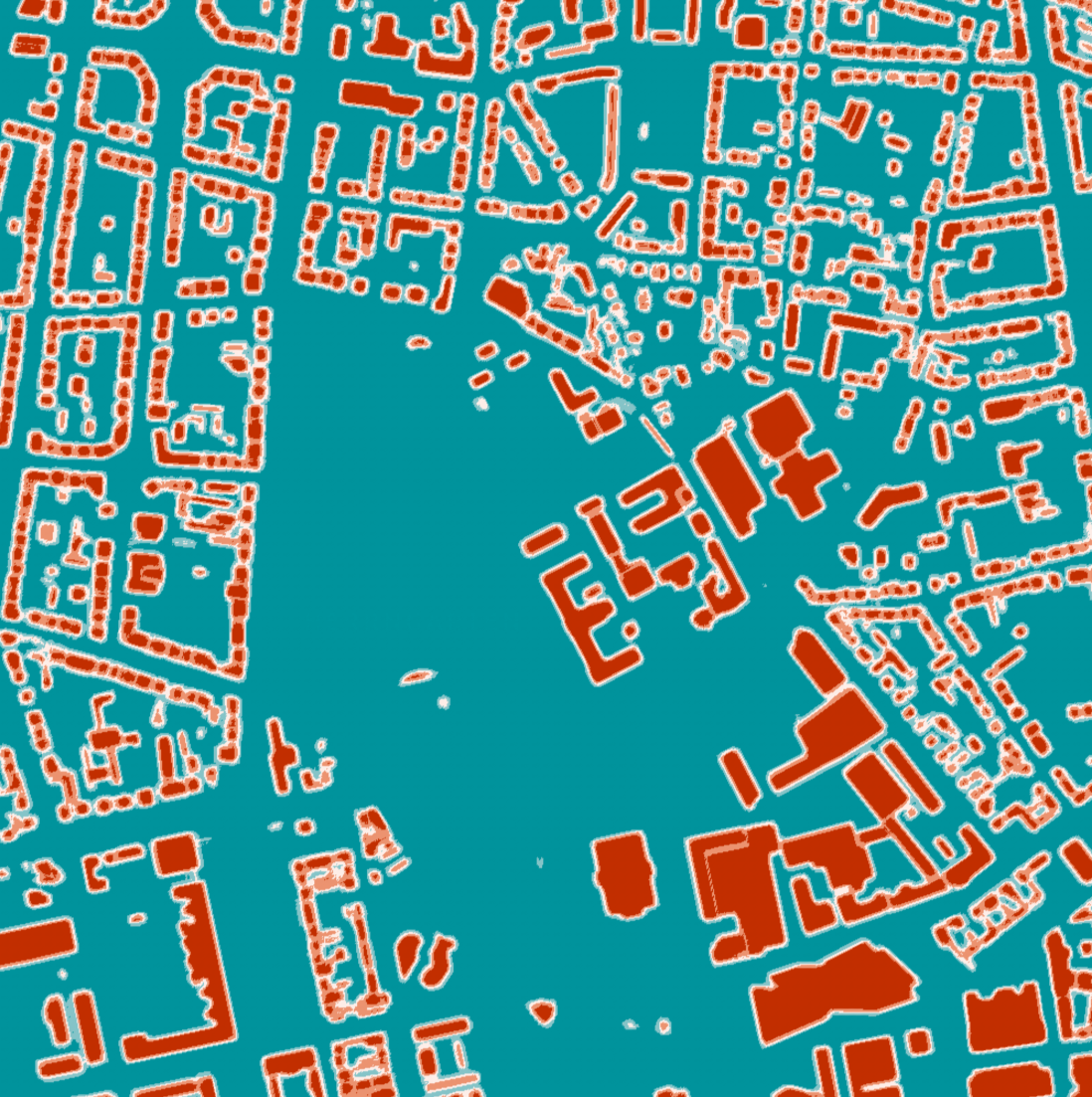}
    \end{subfigure} 
    \begin{subfigure}{0.39\linewidth}
        \includegraphics[width=\textwidth]{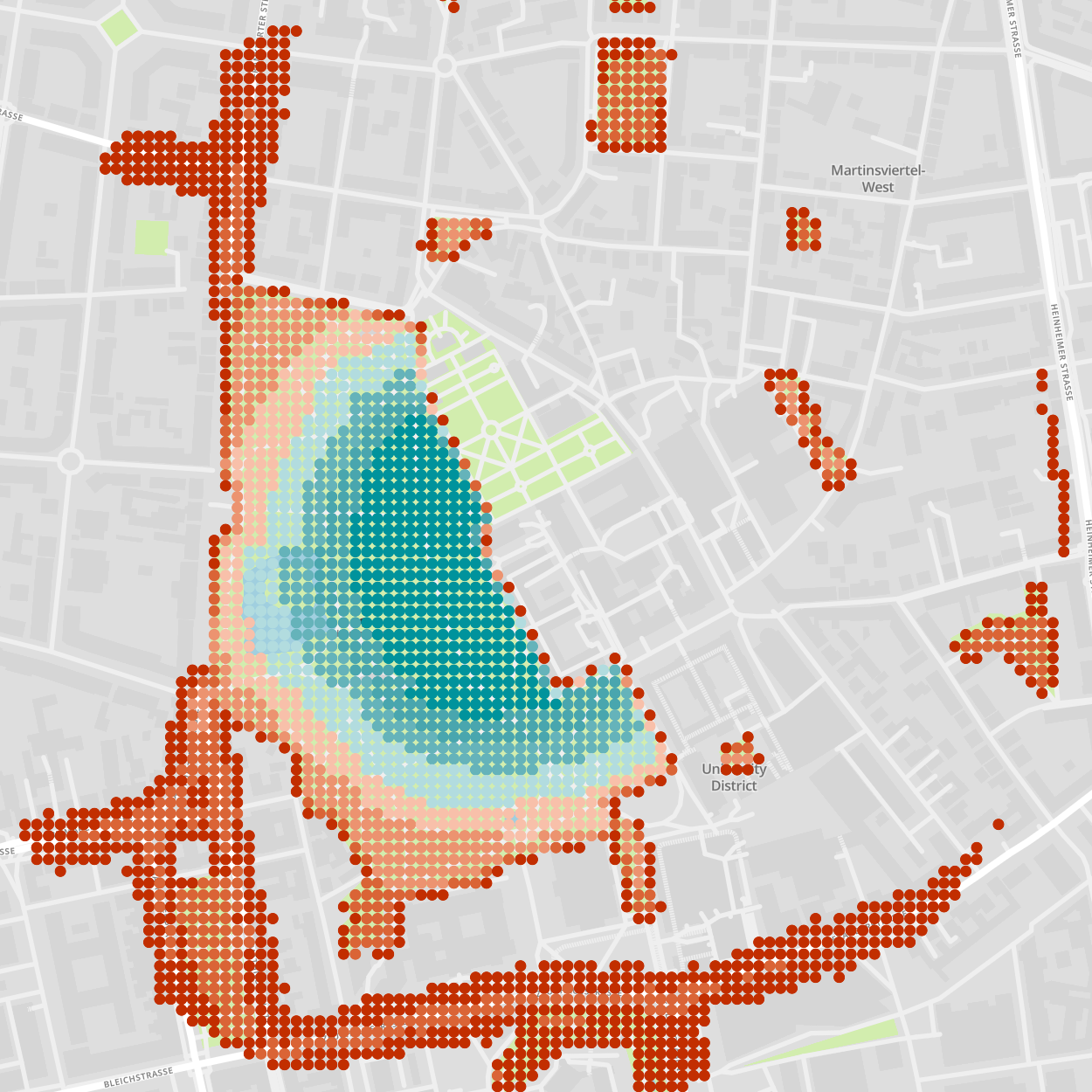}
    \end{subfigure} 
    \caption{
        \textbf{Hybrid relational parameters and Probabilistic Mission Landscape:} 
        Mean (top-left) and variance (top-right) of distance to nearest road and probability of occupation (bottom-left) parameterize the hybrid relational, logical model for defining a Probabilistic Mission Landscape (bottom-right).
    }
    \label{fig:pml_matrix}
\end{figure}

%% file: listings/promis.tex
\begin{listing}
    \centering
    \begin{minted}
    [
        frame=none,
        autogobble,
        fontsize=\footnotesize,
    ]{prolog}
    % Distributional clauses of distances in meters
    distance(x0, y0, building) ~ normal(20, 0.5).
    ...
    
    % Probabilistic facts describing occupancy 
    0.9::over(x0, y0, primary).
    ...
        
    % Mission parameters
    1.0::standard; 0.0::special.
    1.0::day; 0.0::night.
    
    % Weather
    1/10::fog; 9/10::clear.
    
    % Visual line of sight
    vlos(X, Y) :-
        night, clear, distance(X, Y, operator) < 50;
        day, fog, distance(X, Y, operator) < 250;
        day, clear, distance(X, Y, operator) < 500.
    
    % City rules to fly over parks or roads
    local_rules(X, Y) :-
        over(X, Y, park); 
        distance(X, Y, primary) < 15;
        distance(X, Y, secondary) < 10;
        distance(X, Y, tertiary) < 5.
    
    % The Probabilistic Mission Landscape
    landscape(X, Y) :- 
        standard, local_rules(X, Y), vlos(X, Y);
        special, local_rules(X, Y), day;
        special, local_rules(X, Y), night, vlos(X, Y).
    \end{minted}
    \caption{
        \textbf{Probabilistic legal- and safety-verification:}
        While the operator usually has to decide on-site whether and how they can maneuver their UAV, ProMis models and automates this decision.
        Probabilistic inference over the navigation space utilizing this model and extracted distributional knowledge of the environment yields a PML for mission design.
    }
    \label{listing:uam_model}
\end{listing}

%% file: figures/architecture.tex
\begin{figure*}
    \centering
    \includegraphics[width=0.7\textwidth]{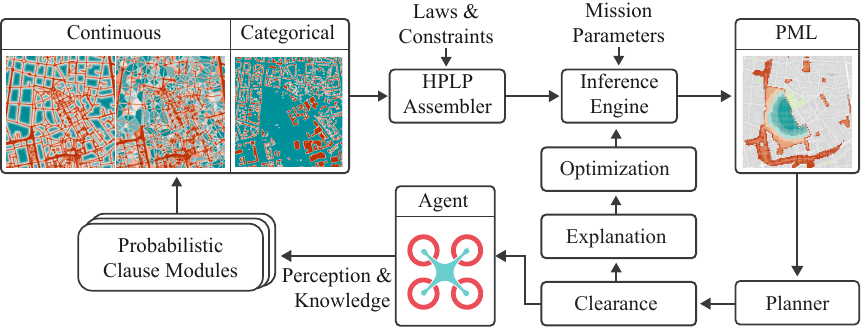}
    \caption{
        \textbf{The Probabilistic UAS architecture:}
        Perception and knowledge of the agent are mapped into continuous and categorical distributions through the Probabilistic Clause Modules (PCM).
        These parameters are then, together with predefined laws, constraints and operator preferences, assembled as distributional clauses into a complete Hybrid Probabilistic Logic Program (HPLP).
        The Probabilistic Mission Landscape, obtained via inference on the resulting HPLP, is then used as basis for laying out an initial mission plan.
        If the proposed plan is denied clearance, a model explanations indicates critical parameters to inform an optimizer on generating a valid route.
        Otherwise, the planned mission can commence.
    }
    \label{fig:architecture}
\end{figure*}

%% file: figures/ceo.tex
\begin{figure*}
    \centering
    \begin{subfigure}{0.31\textwidth}
        \includegraphics[width=\textwidth]{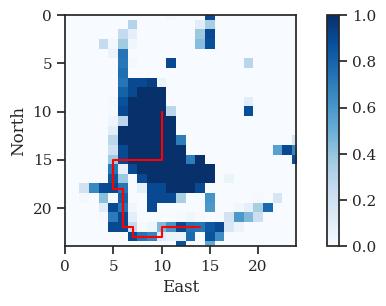}
        \caption{}
    \end{subfigure}
    \begin{subfigure}{0.33\textwidth}
        \includegraphics[width=\textwidth]{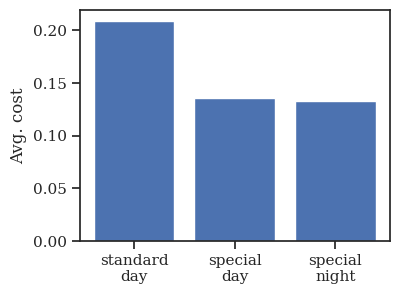}
        \caption{}
    \end{subfigure}
    \begin{subfigure}{0.31\textwidth}
        \includegraphics[width=\textwidth]{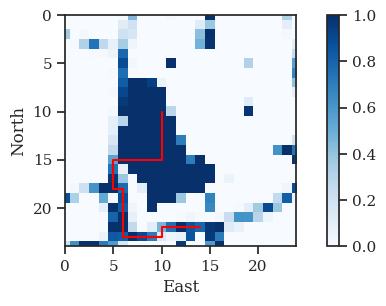}
        \caption{}
    \end{subfigure}
    \caption{
        \textbf{Probabilistic Clearance, Explanation and Optimization results:}
        Once a PML is constructed, the journey can be established through spaces likely to comply with the employed HPLP.
        Then, the clearance check validates the proposed trajectory against the PML by considering its normalized cost, which can induce rejection of the journey due to its violations of navigation constraints.
        Picture (a) shows such a scenario in which a proposed trajectory using a restrictive \textit{standard} license at \textit{day} is denied clearance.
        As a result, one can employ explanation (b) to get an overview of the impact of changes in the mission plan, e.g., changing to a more permissive license for \textit{special} operations. 
        Note that using a \textit{standard} license at \textit{night} does not entail any valid path; hence its cost is not shown in (b).
        While the PML seen in (c) looks similar to (a), the change in mission parameters has raised the probabilities along the optimal path and is granted clearance.
    }
    \label{fig:ceo}
\end{figure*}

%% file: listings/overpass.tex
\begin{listing}[t]
    \centering
    \begin{minted}
    [
        frame=none,
        autogobble,
        fontsize=\footnotesize
    ]{c}
    // Output format and timeout for request
    [out:json][timeout:25];

    // Requested features within a bounding box
    (
        way["highway"="primary"]({{bbox}});
        ...
    );

    // Return retrieved data
    out body; >; out skel qt;
    \end{minted}
    \caption{
        \textbf{Querying geospatial data from OpenStreetMap:}
        Analogously to the HPLP in Listing~\ref{listing:uam_model}, OpenStreetMap uses relational information that describes the types of nodes, ways and areas.
        Hence, we can utilize the same geospatial tags within the HPLP's spatial relations \textit{distance} and \textit{over}. 
    }
    \label{listing:overpass_ql}
\end{listing}

%% file: content/results.tex
\section{Results}
\label{sec:results}

We now present results on applying a CEO cycle on top of the ProMis framework to (i) propose an initial path over an urban environment using operator-specific rules, (ii) explain which parameter denied the initial proposal its clearance, and (iii) obtain the optimal settings and path to traverse the PML.
We employ two distributional clauses, \textit{distance}  and \textit{over}, derived from geospatial data queried from OpenStreetMap and represented as shown in Listing~\ref{listing:overpass_ql}.
Further, we assume for all map features to be distributed according to a Gaussian translation $\vec{t} \sim \mathcal{N}(0, \text{diag}(\qty{10}{\metre}, \qty{10}{\metre}))$ on their origin, representing the uncertainty of crowd-sourced map data.
No further disturbances on the data are assumed, e.g., no rotational or scaling effects impact the samples.

Figure~\ref{fig:ceo} showcases the CEO cycle on the model presented in Listing~\ref{listing:uam_model}.
Since the journey traverses an urban environment, various streets, buildings, and parks are involved in the PML's definition.
The UAS must traverse a park, starting at its northeastern border, aiming to reach its goal location in the south.
We have chosen a mission space of $\qty{1}{\square\kilo\metre}$ represented by a PML that was sampled in a $25 \times 25$ grid.

The setting has two mission parameters which influence the entire PML:
First, the pilot's license can be either \textit{standard} or \textit{special}, representing, e.g., whether they completed a simple online questionnaire or a more involved test, including flight hours.
Second, the mission can ensue either day or night, influencing, for example, the visual line of sight between the operator and UAS.
Initially, employing a standard license during the day, the best route is denied clearance.
Therefore, the system explains to the operator how mission settings influence the result individually.
Subsequently, one can identify the optimal mission setting to reach the goal without violating the modeled rules.

%% file: content/conclusion.tex
\section{Conclusion}
\label{sec:conclusion}

This paper presents how to realize Clearance, Explanation, and Optimization in a probabilistic mission design framework such as ProMis.
Thus, we have taken steps towards safe and reliable UAS navigation that both represents constraints in a human-interpretable, declarative language and incorporates uncertainties within the underlying data and sensors.
While prior work has unified geospatial data, operator rules, and legislation into a hybrid probabilistic, declarative language, we illustrate its application in navigation to obtain law-compliant routes.
Our presented methods' declarative nature and probabilistic semantics open up an inherently more adaptable and explainable foundation for UAS navigation.
An Open Source implementation accompanies our contribution to further progress development on such probabilistic mission design systems.

Future work leads to tackling inherently dynamic scenarios, including asynchronous parameter updates and dealing with multi-agent systems.
Similarly, integrating Deep Learning methods while using robust learning architectures within this navigation architecture is a promising direction.
Furthermore, demonstrating real-world applications and overcoming hardware limitations to enable computations on the agent rather than centrally will be critical steps in underlining the advantages of ProMis and the CEO cycle.
Finally, beyond its application in UAS navigation, general robotics applications can also significantly benefit from these advancements.

\section*{Acknowledgments}

The Eindhoven University of Technology authors received support from their Department of Mathematics and Computer Science and the Eindhoven Artificial Intelligence Systems Institute.
Map data \copyright~OpenStreetMap contributors, licensed under the Open Database License (ODbL) and available from https://www.openstreetmap.org.
Map styles \copyright~Mapbox, licensed under the Creative Commons Attribution 3.0 License (CC BY 3.0) and available from https://github.com/mapbox/mapbox-gl-styles.

%% file: main.bbl
\begin{thebibliography}{10}
\providecommand{\url}[1]{#1}
\csname url@samestyle\endcsname
\providecommand{\newblock}{\relax}
\providecommand{\bibinfo}[2]{#2}
\providecommand{\BIBentrySTDinterwordspacing}{\spaceskip=0pt\relax}
\providecommand{\BIBentryALTinterwordstretchfactor}{4}
\providecommand{\BIBentryALTinterwordspacing}{\spaceskip=\fontdimen2\font plus
\BIBentryALTinterwordstretchfactor\fontdimen3\font minus
  \fontdimen4\font\relax}
\providecommand{\BIBforeignlanguage}[2]{{%
\expandafter\ifx\csname l@#1\endcsname\relax
\typeout{** WARNING: IEEEtran.bst: No hyphenation pattern has been}%
\typeout{** loaded for the language `#1'. Using the pattern for}%
\typeout{** the default language instead.}%
\else
\language=\csname l@#1\endcsname
\fi
#2}}
\providecommand{\BIBdecl}{\relax}
\BIBdecl

\bibitem{Huang2021}
H.~Huang, A.~V. Savkin, and C.~Huang, ``{Drone Routing in a Time-Dependent
  Network: Toward Low-Cost and Large-Range Parcel Delivery},'' \emph{IEEE
  Transactions on Industrial Informatics}, vol.~17, no.~2, pp. 1526--1534, feb
  2021.

\bibitem{Jayalath2021}
K.~Jayalath and S.~R. Munasinghe, ``{Drone-based Autonomous Human
  Identification for Search and Rescue Missions in Real-time},'' in \emph{2021
  10th International Conference on Information and Automation for
  Sustainability (ICIAfS)}.\hskip 1em plus 0.5em minus 0.4em\relax IEEE, aug
  2021, pp. 518--523.

\bibitem{Maddikunta2021}
P.~K. {Reddy Maddikunta}, S.~Hakak, M.~Alazab, S.~Bhattacharya, T.~R.
  Gadekallu, W.~Z. Khan, and Q.~V. Pham, ``{Unmanned Aerial Vehicles in Smart
  Agriculture: Applications, Requirements, and Challenges},'' \emph{IEEE
  Sensors Journal}, vol.~21, no.~16, pp. 17\,608--17\,619, 2021.

\bibitem{Wu2020}
C.~Wu, B.~Ju, Y.~Wu, and N.~Xiong, ``{SlimRGBD: A Geographic Information
  Photography Noise Reduction System for Aerial Remote Sensing},'' \emph{IEEE
  Access}, vol.~8, no.~3, pp. 15\,144--15\,158, 2020.

\bibitem{Kohaut2023}
S.~Kohaut, B.~Flade, D.~S. Dhami, J.~Eggert, and K.~Kersting, ``{Mission Design
  for Unmanned Aerial Vehicles using Hybrid Probabilistic Logic Programs},'' in
  \emph{26th International Conference on Intelligent Transportation Systems
  (ITSC)}.\hskip 1em plus 0.5em minus 0.4em\relax IEEE, 2023.

\bibitem{problog}
L.~De~Raedt, A.~Kimmig, and H.~Toivonen, ``{Problog: A probabilistic prolog and
  its application in link discovery.}'' in \emph{IJCAI}, vol.~7.\hskip 1em plus
  0.5em minus 0.4em\relax Hyderabad, 2007, pp. 2462--2467.

\bibitem{inference_in_plp}
D.~Fierens, G.~Van~den Broeck, J.~Renkens, D.~Shterionov, B.~Gutmann, I.~Thon,
  G.~Janssens, and L.~De~Raedt, ``{Inference and learning in probabilistic
  logic programs using weighted boolean formulas},'' \emph{Theory and Practice
  of Logic Programming}, vol.~15, no.~3, pp. 358--401, 2015.

\bibitem{undertaking2016european}
{SESAR Joint Undertaking}, ``{European ATM master plan: The Roadmap for
  Delivering High Performing Aviation for Europe: Executive view: Edition
  2015},'' 2015.

\bibitem{sesar_2017}
------, ``{European Drones Outlook Study},'' 2017.

\bibitem{Ec2019}
{The European Commission}, ``{Commission Implementing regulation EU 2019/947 of
  24 May 2019},'' 2019.

\bibitem{Easa2019}
{European Union Aviation Safety Agency}, ``{Acceptable Means of Compliance
  (AMC) and Guidance Material (GM) to Commission Implementing Regulation (EU)
  2019/947},'' pp. 1--130, 2019.

\bibitem{RothwellPatzek2019}
C.~D. Rothwell and M.~J. Patzek, ``{An Interface for Verification and
  Validation of Unmanned Systems Mission Planning: Communicating Mission
  Objectives and Constraints},'' \emph{IEEE Transactions on Human-Machine
  Systems}, vol.~49, no.~6, pp. 642--651, dec 2019.

\bibitem{Rakotonarivo2022}
B.~Rakotonarivo, N.~Drougard, S.~Conversy, and J.~Garcia, ``{Supporting drone
  mission planning and risk assessment with interactive representations of
  operational parameters},'' in \emph{2022 International Conference on Unmanned
  Aircraft Systems (ICUAS)}.\hskip 1em plus 0.5em minus 0.4em\relax IEEE, jun
  2022, pp. 1091--1100.

\bibitem{Primatesta2020}
S.~Primatesta, A.~Rizzo, and A.~la~Cour-Harbo, ``{Ground Risk Map for Unmanned
  Aircraft in Urban Environments},'' \emph{Journal of Intelligent \& Robotic
  Systems}, vol.~97, no. 3-4, pp. 489--509, mar 2020.

\bibitem{Raballand2021}
N.~Raballand, S.~Bertrand, S.~Lala, and B.~Levasseur, ``{DROSERA: A DROne
  Simulation Environment for Risk Assessment},'' in \emph{Proceedings of the
  31st European Safety and Reliability Conference (ESREL 2021)}.\hskip 1em plus
  0.5em minus 0.4em\relax Singapore: Research Publishing Services, 2021, pp.
  354--361.

\bibitem{An2023}
D.~An, R.~Krzysiak, D.~Hollenbeck, and Y.~Chen, ``{Battery-health-aware UAV
  mission planning using a cognitive battery management system},'' in
  \emph{2023 International Conference on Unmanned Aircraft Systems
  (ICUAS)}.\hskip 1em plus 0.5em minus 0.4em\relax IEEE, jun 2023, pp.
  523--528.

\bibitem{Schuchardt2022}
B.~I. Schuchardt, T.~Dautermann, A.~Donkels, T.~J. Lieb, F.~Morscheck,
  M.~Rudolph, and G.~Schwoch, ``{Mission Management and Landing Assistance for
  an Unmanned Rotorcraft for Maritime Operations},'' in \emph{2022 IEEE/AIAA
  41st Digital Avionics Systems Conference (DASC)}, vol. 2022-Septe.\hskip 1em
  plus 0.5em minus 0.4em\relax IEEE, sep 2022, pp. 1--9.

\bibitem{Hohmann2022}
N.~Hohmann, M.~Bujny, J.~Adamy, and M.~Olhofer, ``{Multi-objective 3D Path
  Planning for UAVs in Large-Scale Urban Scenarios},'' in \emph{2022 IEEE
  Congress on Evolutionary Computation (CEC)}.\hskip 1em plus 0.5em minus
  0.4em\relax IEEE, jul 2022, pp. 1--8.

\bibitem{Shen2022}
K.~Shen, R.~Shivgan, J.~Medina, Z.~Dong, and R.~Rojas-Cessa, ``{Multidepot
  Drone Path Planning With Collision Avoidance},'' \emph{IEEE Internet of
  Things Journal}, vol.~9, no.~17, pp. 16\,297--16\,307, sep 2022.

\bibitem{Cabral2023}
K.~Cabral, J.~Silveira, C.-A. Rabbath, and S.~Givigi, ``{Hierarchical
  Cooperative Assignment Algorithm (CAA) for mission and path planning of
  multiple fixed-wing UAVs based on maximum independent sets},'' in \emph{2023
  International Conference on Unmanned Aircraft Systems (ICUAS)}.\hskip 1em
  plus 0.5em minus 0.4em\relax IEEE, jun 2023, pp. 695--702.

\bibitem{Papaioannou2021}
S.~Papaioannou, P.~Kolios, T.~Theocharides, C.~G. Panayiotou, and M.~M.
  Polycarpou, ``{3D Trajectory Planning for UAV-based Search Missions: An
  Integrated Assessment and Search Planning Approach},'' in \emph{2021
  International Conference on Unmanned Aircraft Systems (ICUAS)}.\hskip 1em
  plus 0.5em minus 0.4em\relax IEEE, jun 2021, pp. 517--526.

\bibitem{Hu2020}
X.~Hu, B.~Pang, F.~Dai, and K.~H. Low, ``{Risk Assessment Model for UAV
  Cost-Effective Path Planning in Urban Environments},'' \emph{IEEE Access},
  vol.~8, pp. 150\,162--150\,173, 2020.

\bibitem{nittihybrid}
D.~Nitti, T.~De~Laet, and L.~De~Raedt, ``{Probabilistic logic programming for
  hybrid relational domains},'' \emph{Machine Learning}, vol. 103, no.~3, pp.
  407--449, 2016.

\bibitem{kumar2022first}
N.~Kumar, O.~Kuzelka, and L.~De~Raedt, ``{First-Order Context-Specific
  Likelihood Weighting in Hybrid Probabilistic Logic Programs},'' \emph{arXiv
  preprint arXiv:2201.11165}, 2022.

\bibitem{flade2021error}
B.~Flade, S.~Kohaut, and J.~Eggert, ``{Error Decomposition for Hybrid
  Localization Systems},'' in \emph{2021 IEEE International Intelligent
  Transportation Systems Conference (ITSC)}.\hskip 1em plus 0.5em minus
  0.4em\relax IEEE, 2021, pp. 149--156.

\end{thebibliography}
